\pdfoutput=1
\documentclass[11pt,onecolumn]{cleantechnicalreport}

\usepackage[authoryear,sort&compress,round]{natbib}
\usepackage{hybridfrontmatter}
\usepackage{multirow}
\usepackage{algorithm}
\usepackage{algorithmic}
\usepackage{float}

\renewcommand{\ReportTitleFont}{%
  \normalfont\bfseries\fontsize{14}{17}\selectfont}

\hypersetup{
  colorlinks=true,
  linkcolor=blue,
  citecolor=blue,
  urlcolor=blue,
  pdftitle={SKT: Skill-Use Training at Scale via Verified Synthetic Data Generation},
  pdfauthor={Zelin Tan, Yiqun Zhang, Hao Li, Zhiyao Cui, Hejia Geng,
    Shao Zhang, Hangfan Zhang, Yang Chen, Xiaosong Wang, Lilong Wang,
    Zhenfei Yin, Chen Zhang, Shuyue Hu, Lei Bai}
}

\makeatletter
\fancypagestyle{skillsbenchfooter}{%
  \fancyhf{}%
  \fancyhead[C]{\footerfont\@title}%
  \fancyfoot[L]{\footerfont
    \href{https://openreward.ai/benchflow/skillsbench}{%
      \nolinkurl{https://openreward.ai/benchflow/skillsbench}}}%
  \fancyfoot[R]{\footerfont\thepage}%
  \renewcommand{\headrule}{\color{gray}\DefaultHeadRule}%
  \renewcommand{\footrule}{\textcolor{gray}{\DefaultFootRule}}%
}
\makeatother

\let\cite\citep
\newcommand{\method}{\mbox{SKT}}
\title{SKT: Skill-Use Training at Scale via Verified Synthetic Data Generation}
\author{Zelin~Tan\ReportAuthorMark{$\dagger$}}
\author{Yiqun~Zhang}
\author{Hao~Li}
\author{Zhiyao~Cui}
\author{Hejia~Geng}
\author{Shao~Zhang}
\author{Hangfan~Zhang}
\author{Yang~Chen}
\author{Xiaosong~Wang}
\author{Lilong~Wang}
\author{Zhenfei~Yin}
\author{Shuyue~Hu\ReportAuthorMark{$\ddagger$}}
\author{Chen~Zhang\ReportAuthorMark{$\ast$}}
\author{Lei~Bai\ReportAuthorMark{$\ddagger \ast$}}
\affil{\textbf{Shanghai Artificial Intelligence Laboratory}}

\reporthuggingfacetrain{https://huggingface.co/datasets/Artemis0430/deepseek-v4-pro-opencode-skill-use-data-2487}
\reporthuggingfaceeval{https://huggingface.co/datasets/Artemis0430/skilleval-v1}
\reportcontactdisplay{zhangchen1@pjlab.org.cn,tanzl@mail.ustc.edu.cn}{\textbf{\{zhangchen1,bailei\}@pjlab.org.cn}}
\reportauthornote{$^{\ast}$ Corresponding authors.\quad $^{\ddagger}$ Project Lead.}
\reportfootertext{$^{\dagger}$ Work done during an internship at Shanghai Artificial Intelligence Laboratory.}

\begin{abstract}
Agent Skills have become an important mechanism for equipping language-model
agents with reusable procedural knowledge. However, providing skills alone
does not guarantee that current models can identify, apply, and coordinate
them effectively. To improve models' skill-use capabilities, we introduce
SKT, a verified data synthesis pipeline that constructs
skill-grounded tasks and executable trajectories from large collections of
Agent Skills. SKT selects suitable single-skill and multi-skill
configurations, synthesizes tasks through rule-based and agent-based verification
with feedback-guided repair, and retains only successful trajectories that
substantively use every required skill. Using 2,000 public skills,
SKT produces 4,000 task packages and 27,164 verified trajectories.
Based on the same pipeline and a disjoint task pool, we further construct
SkillEval, a held-out executable benchmark for evaluating skill use.
Experiments across diverse models, benchmarks, and agent harnesses show that
supervised fine-tuning on SKT-generated trajectories consistently
improves skill-use performance. Verification ablations, cross-harness
evaluation, and scaling experiments further show that these gains depend on
high-quality supervision, extend beyond a single agent interface, and increase
with broader skill coverage. Together, these results establish verified data
synthesis as an effective and scalable approach to skill-use training.
\end{abstract}

\begin{document}

\maketitle

\providecommand{\method}{\textsc{SKT}}

\section{Introduction}

An \emph{Agent Skill} is an organized package of instructions, metadata, and optional resources (e.g., scripts and templates) that can be discovered, loaded, and utilized by a large language model (LLM) \citep{zhang2025agentskills}. By encapsulating domain knowledge, procedural workflows, and auxiliary tools, Agent Skills extend model capabilities beyond their intrinsic knowledge, adapt them to domain-specific tasks, facilitate the sharing and reuse of expert knowledge, and support the composition of multiple skills for solving complex tasks. According to public records, since October 2025, more than 600,000 skills have been made publicly available, and this number continues to grow \citep{zhang2025agentskills,qu2026skills}.

However, the availability of a skill does not necessarily imply that a model can effectively utilize it. Although recent studies have demonstrated that proper skill utilization can substantially improve task success rates, a great number of LLMs do not inherently possess this ability \citep{li2026skillsbench,han2026sweskillsbench}. Effective skill utilization remains a non-trivial challenge: a model must be able to identify a skill when it is relevant and applicable, understand and follow its operational constraints, coordinate it with other complementary skills, and correctly execute the prescribed procedures following skill instructions while effectively utilizing the associated tools \citep{li2026agentskillos}.

In this paper, we investigate how to enable LLMs to effectively utilize skills. Despite the growing interest in Agent Skills, this fundamental question remains largely underexplored. Existing studies have primarily focused on efficiently retrieving skills from skill ecosystems \citep{li2026agentskillos}, internalizing skills into models to reduce reliance on external skill documents \citep{lu2026skill0,lin2026skillc,zhu2026skill05,he2026siri}, and leveraging skills to support the self-improvement of LLM-based agents \citep{wang2025sage,xia2026skillrl,shi2026skill1,vishe2026skillr1}. However, these approaches mainly address how skills are retrieved, represented, or incorporated into models, rather than how models can effectively understand, coordinate, and leverage skills to solve diverse tasks.

To this end, we introduce Skill-use Training (SKT), a framework for synthesizing high-quality training data to equip LLMs with effective skill-use capabilities. The key idea of SKT is to construct skill-grounded tasks that are solvable, verifiable, and complexity-controllable, enabling models to learn when and how to apply skills across diverse scenarios. SKT first curates skills from public skill repositories \citep{qu2026skills} and constructs a candidate skill pool based on quality, diversity, and composability. Based on this pool, SKT synthesizes skill-grounded tasks with controlled complexity, ranging from single-skill scenarios to multi-skill compositions. Each task includes not only structured specifications but also reference solutions and verification mechanisms, ensuring task solvability and verifiability. These tasks later undergo automatic assessment to ensure appropriate difficulty and genuine dependence on the designated skills. Finally, SKT collects and filters execution trajectories from a strong LLM through an interaction harness, retaining only successful trajectories with faithful skill utilization as high-quality supervision for skill-use training.

\begin{figure}[b]
    \centering
    \includegraphics[width=0.98\textwidth]{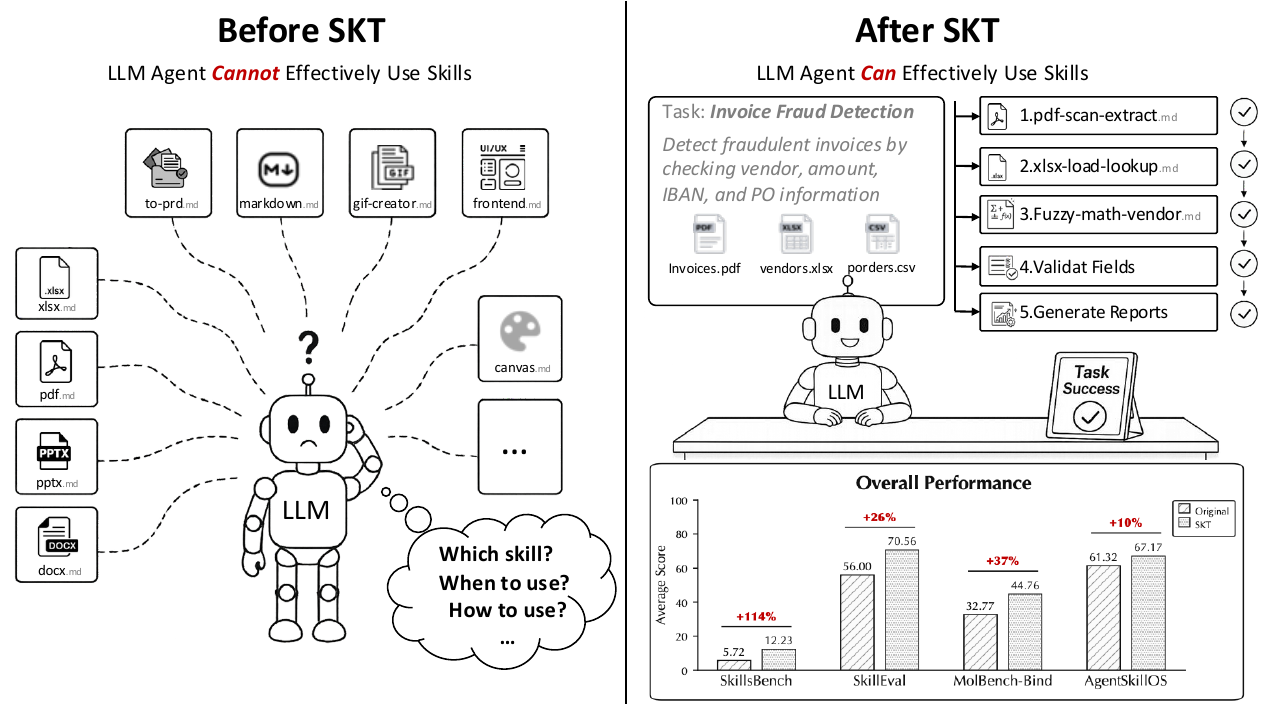}
    \caption{SKT is a multi-agent-based data synthesis pipeline for skill-use
    tasks and trajectories. Training on data synthesized by SKT can effectively
    improve agents' ability to use Agent Skills.}
    \label{fig:skt_overview}
\end{figure}


Using SKT, we construct a skill-use training corpus from 2,000 public skills, producing 4,000 synthetic tasks and 27,164 verified execution trajectories. We use these trajectories as supervision to train Qwen3.5-9B \cite{qwen2026qwen35} and Gemma~4 E4B-IT \cite{gemmateam2026gemma4}, and evaluate the resulting models on four benchmarks: SkillsBench \cite{li2026skillsbench}, MolBench-Bind \cite{zhang2026molclaw}, AgentSkillOS-bench \cite{li2026agentskillos} , and SkillEval. The latter is constructed using SKT from a held-out skill pool with no overlap with the training corpus, demonstrating that SKT can also support scalable benchmark construction. As shown in Figure~\ref{fig:skt_overview}, models trained with SKT consistently achieve improvements in skill-use performance across different harnesses and benchmarks. Further analyses, including ablations of verification components, cross-harness evaluations, and scaling studies, demonstrate that these gains arise from high-quality supervision, generalize across agent interfaces, and scale with broader skill coverage. 


Our key contributions are threefold:

\begin{itemize}
    \item We introduce \method{}, a framework that empowers LLMs with the ability to effectively utilize Agent Skills.
    \item We develop a scalable data synthesis pipeline that generates skill-grounded training tasks with solvability, verifiability, and complexity control, and supports the construction of held-out benchmarks.
    \item We present SkillEval, a new benchmark for skill-use evaluation, and conduct extensive experiments across models, harnesses, and benchmarks, demonstrating the effectiveness, scalability, and generalizability of SKT.
\end{itemize}

\section{Related Work}

\paragraph{Learning and evolving external skills.}
A line of work treats skills as persistent, editable memories that co-evolve
with an agent. SAGE accumulates executable skills across sequential rollouts
and rewards both skill creation and reuse \citep{wang2025sage}. SkillRL
distills successful demonstrations and failure lessons into a hierarchical
SkillBank, then updates the bank from validation failures as the policy learns
\citep{xia2026skillrl}. Skill1 more tightly couples the lifecycle by training
one policy to select, use, and distill skills from a shared outcome signal
\citep{shi2026skill1}. Skill-R1 instead freezes the task model and trains a
lightweight editor to revise textual skills over multiple verifier-scored
generations \citep{vishe2026skillr1}. 

\paragraph{Skill internalization.}
Another line uses external skills as temporary training scaffolds. SKILL0
progressively withdraws skills according to their measured on-policy
helpfulness \citep{lu2026skill0}, whereas SkillC turns paired rollouts with and
without skills into a direct contrastive credit-assignment signal
\citep{lin2026skillc}. SIRI mines skills from the policy's own successful
rollouts, validates their utility, and distills only beneficial skill-guided
actions into a skill-free policy \citep{he2026siri}. Skill0.5 adopts a hybrid
strategy, internalizing general skills while enforcing the use of
task-specific skills to improve out-of-distribution transfer
\citep{zhu2026skill05}.

\paragraph{Parameterized skill representations.}
Weight-space approaches replace repeated skill-text injection with modular
parameters. Skill-to-LoRA synthesizes skill-guided demonstrations and
distills each \texttt{SKILL.md} into a separately loadable adapter
\citep{zhang2026skilltolora}. LatentSkill and ParametricSkills instead train
hypernetworks that map textual skills to LoRA weights without per-skill
backbone tuning; both investigate parameter-space composition, while
ParametricSkills also trains on single- and multi-turn skill-exploitation
trajectories and supports test-time skill evolution
\citep{yu2026latentskill,zhao2026parametricskills}. 

\paragraph{Evaluating Agent Skills.}
SkillsBench measures the marginal value of curated skills through paired
skill/no-skill execution across diverse domains and shows that focused skill
sets can outperform exhaustive bundles \citep{li2026skillsbench}.
SWE-Skills-Bench applies controlled, execution-based evaluation to real
software repositories and finds that gains are highly dependent on skill
specificity and contextual compatibility \citep{han2026sweskillsbench}.
AgentSkillOS studies skill discovery and DAG-based orchestration at ecosystem
scale, together with an artifact-production benchmark
\citep{li2026agentskillos}; MolClaw organizes drug-discovery tools into a
three-level skill hierarchy and introduces MolBench to evaluate the resulting
agent on long-horizon workflows \citep{zhang2026molclaw}. 

\paragraph{Agentic data synthesis.}
STEPS organizes capabilities into a
hierarchical skill taxonomy and samples coherent skill combinations for
compositional instruction synthesis \citep{wei2026steps}. AgentSynth composes
independently generated subtasks into long-horizon computer-use tasks, with
complexity controlled by the number of subtasks \citep{xie2026agentsynth}.
For terminal agents, TermiGen jointly synthesizes executable environments and
trajectories containing error-recovery behavior \citep{zhu2026termigen}, while
CLI-Universe constructs and verifies containerized tasks sampled from a
multidimensional capability taxonomy \citep{hua2026cliuniverse}. SkillSynth
samples workflows from a scenario-mediated skill graph to diversify required
execution paths \citep{fan2026skillsynth}. Terminal-World directly uses Agent
Skills to co-generate task instructions, environments, and teacher
trajectories, and composes skills into teams and graphs for broader task
coverage \citep{cheng2026terminalworld}.
\section{Method}\label{sec_method}

\begin{figure}[b]
    \centering
    \includegraphics[width=\textwidth]{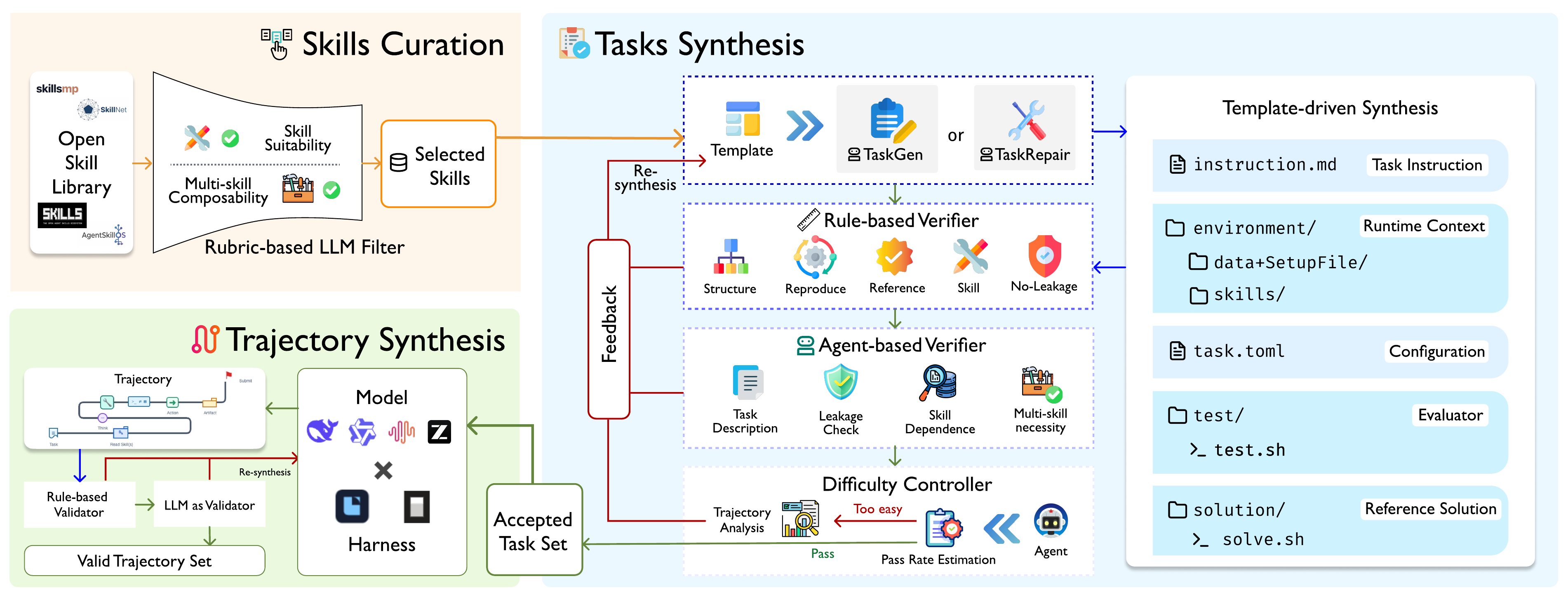}
    \caption{Overview of \method{}. A rubric-based filter selects suitable
    single- and multi-skill configurations. Template-driven synthesis produces
    executable task packages, which pass through rule-based, agent-based, and
    difficulty checks; failed tasks return to the author through a feedback
    loop. Accepted tasks are then solved by a model--harness pair, and a second
    verification stage retains only correct and faithful skill-use
    trajectories for SFT; rejected rollouts are freshly resampled without
    verifier feedback. The right panel expands the task-package template.}
    \label{fig:method_overview}
\end{figure}

This section presents our three-stage pipeline \method{}. It first selects suitable
single- and multi-skill configurations, then synthesizes and verifies
executable task packages, and finally collects and validates skill-use
trajectories. We also describe how the verified trajectories are used for
supervised fine-tuning. Figure~\ref{fig:method_overview} summarizes the overall
process.

\subsection{Stage I: Skill Curation}\label{subsec_skill_selection}

From a candidate collection $\mathcal{C}$, a rubric-based LLM judge retains
skills that can ground executable, objectively checkable tasks. We then sample
a skill set $\mathcal{S} \subseteq \mathcal{C}$ with configurable cardinality
$k = |\mathcal{S}|$. For $k > 1$, a composition judge retains $\mathcal{S}$
only if its skills form a coherent workflow with distinct roles; rejected sets
are resampled. For $k = 1$, skills are sampled directly. Our experiments use
$k \in \{1,2,3\}$. The selected skills are then passed to the task synthesis
stage.

\subsection{Stage II: Task Synthesis}
\label{subsec_task_synthesis}

\paragraph{Template-driven authoring.}
Given \(\mathcal{S}\), \textsc{TaskGen} reads every skill and fills the fixed
template in Figure~\ref{fig:method_overview}. The task package \(\mathcal{T}\)
contains the instruction, isolated runtime, execution settings, and executable
evaluator, and reference solution. It must be objectively gradable, and assign an intended role to every selected skill.

\paragraph{Rule-based task verifier.}
We first apply deterministic checks to failures that can be established from
the package itself. The verifier confirms that all required components are
present, file paths are valid, evaluator definitions are supported, and the
authoring trace records access to every skill in \(\mathcal{S}\). It then executes the reference solution in a clean workspace initialized only from the task inputs; the produced artifacts must receive full evaluator credit. Finally, it compares solver-visible task
materials against the references and recorded skill requirements to detect
reference values or skill rules copied verbatim outside the supplied skill
files.

\paragraph{Agent-based task verifier.}
Deterministic checks cannot establish whether a task is semantically well
posed. An LLM verifier therefore reviews the task package and checks that the
instruction clearly specifies the required artifact and output location, all
information needed for completion is available in the local environment, and
solver-visible materials outside the supplied skill files do not semantically
reveal the hidden answer, evaluation logic, or decisive skill-derived
requirements.

\emph{Skill dependence.} The verifier uses paired rollouts to check the task's dependence on skill: \(\mathcal{S}\) is available in one rollout and
withheld in the other, while the task and all remaining settings are fixed.
The evaluator scores both outputs, after which the verifier compares the
score difference to determine whether skill access benefits
task completion at the bundle level. 

\paragraph{Difficulty control.}
To filter tasks that are valid but too easy, a fixed solver and harness attempt
\(\mathcal{T}\) with the selected skills available for \(N\) independent
rollouts. If \(r_j\) is the evaluator score of attempt \(j\), we estimate
\[
p_{\mathrm{pass}}(\mathcal{T},\mathcal{S})
=\frac{1}{N}\sum_{j=1}^{N}\mathbf{1}[r_j=1].
\]
A task with \(p_{\mathrm{pass}}\geq\theta_{\mathrm{easy}}\) is sent back for
repair. The controller analyzes its rollouts to identify shortcuts and suggest
how to increase reasoning or execution demands.

\paragraph{Feedback-guided task repair.}
The first failing gate sends \textsc{TaskRepair} concrete rule errors, semantic
agent feedback, or trajectory-grounded difficulty suggestions. After the
package is revised, validation restarts at the rule-based gate so that new
defects are detected. Only task--configuration pairs that pass all three gates
enter \(\mathcal{D}_{\mathrm{task}}\); candidates that exhaust the repair budget
are discarded.

\subsection{Stage III: Trajectory Synthesis}
\label{subsec_trajectory_synthesis}

For each \((\mathcal{T},\mathcal{S})\in\mathcal{D}_{\mathrm{task}}\) and each
configured teacher--harness pair \((m,h)\), teacher \(m\) solves the task through
harness \(h\) with its files and selected skills available. We retain the
complete trajectory \(\tau\), including model messages, tool calls and results,
and output artifacts.

The rule-based validator requires full evaluator credit, normal and complete
termination, a well-formed tool trace, and explicit skill access. The LLM
validator then checks that each skill was consulted before the action it should
guide, affected concrete decisions or operations, and was applied correctly. A failed rollout
is resampled from the same task without exposing the failure or verifier
feedback to the next attempt. The first rollout that passes both validators
for a task--teacher--harness pairing enters \(\mathcal{D}_{\mathrm{traj}}\);
exhausting its rollout budget yields no training example for that pairing.

\subsection{Training on Verified Trajectories}
\label{subsec_trajectory_supervision}

The valid trajectory set at the output of Figure~\ref{fig:method_overview} is
converted into harness-native training examples. We apply masked
autoregressive supervised fine-tuning; for a tokenized trajectory
\(\tau=(x_1,\ldots,x_T)\), the objective is
\[
\mathcal{L}_{\mathrm{SFT}}
=-\sum_{t\in\mathcal{A}(\tau)}
\log p_{\theta}(x_t\mid x_{<t}),
\]
where \(\mathcal{A}(\tau)\) contains all assistant-generated reasoning,
tool-call, and response tokens. System and task messages and tool observations
remain in the conditioning context but are masked from the loss. Thus the
model learns the complete skill-use process---from locating relevant
instructions to executing and checking skill-guided actions---rather than only
the final answer.

\section{Experiments}\label{sec_experiments}

\subsection{Experimental Setup}\label{subsec_experimental_setup}

\paragraph{Models and Training.}
We study Qwen3.5-9B~\cite{qwen2026qwen35} and Gemma~4
E4B-IT~\cite{gemmateam2026gemma4}, using full-parameter supervised
fine-tuning on the end-to-end trajectory objective from
Section~\ref{sec_method}.
All runs use LLaMA-Factory~\cite{zheng-etal-2024-llamafactory} for one epoch,
with a learning rate of \(5\times10^{-6}\), a cosine schedule, a \(3\%\)
warmup ratio, bfloat16 precision, and an effective batch size of eight. We
retain complete trajectories whose tokenized input targets contain at most 64k tokens.

\paragraph{Data Synthesis.}
From the public \texttt{skills.sh} library~\cite{qu2026skills}, we select 2,000
distinct skills and synthesize 4,000 accepted task packages: 1,520
single-skill, 1,295 two-skill, and 1,185 three-skill tasks. Skill selection,
task generation and repair, skill-dependence probing, and task and trajectory
verification use DeepSeek V4 Pro~\cite{deepseekai2026deepseekv4} with a Claude
Agent harness~\cite{anthropic2025claudeagentsdk}; difficulty control uses
Qwen3.5-35B-A3B with OpenCode~\cite{anomaly2026opencode}. We run five
difficulty rollouts and repair a task if at least three receive full credit
(\(N=5\), \(\theta_{\mathrm{easy}}=0.6\)).

\paragraph{Trajectory Collection.}
We pair four teachers---MiniMax-M2.5~\cite{minimax2026m2series},
GLM-5~\cite{glm5team2026glm5}, Qwen3.5-397B-A17B, and DeepSeek V4 Pro---with
both DeepAgents~\cite{langchain2025deepagents} and OpenCode. From the 32,000
candidate task--teacher--harness combinations, verification retains 14,277
DeepAgents and 12,887 OpenCode trajectories. Each is one complete,
uninterrupted rollout.

\paragraph{Public Benchmarks.}
We evaluate on three public benchmarks. \emph{SkillsBench}%
\thispagestyle{skillsbenchfooter}
\cite{li2026skillsbench} uses the 77-task
\texttt{test} split of the OpenReward
implementation. It evaluates multi-step, skill-based problem solving across
engineering, science, finance, software development, and data processing.
Each task runs in an isolated container and receives a normalized reward from
task-specific tests, which can assign partial credit; no LLM grader is used.
\emph{MolBench-Bind} is the
37-task MS2
binding-affinity subset of MolBench~\cite{zhang2026molclaw}. Each task presents
a protein target and two candidate molecules as SMILES strings and asks the
agent to select the molecule with higher or lower affinity under the specified
$K_i$ direction. \emph{AgentSkillOS-Bench}
\cite{li2026agentskillos} contains 30 multi-format artifact-production tasks,
with six tasks each in data computation, document creation, motion/video,
visual creation, and web interaction.

\paragraph{SkillEval.}
We construct SkillEval using the same task-synthesis pipeline that produces the
tasks underlying our training trajectories, but from a separately generated
pool. Neither these tasks nor trajectories collected on them are included in
the SFT data. SkillEval is a cross-domain executable benchmark of
skill-grounded agent capabilities. It covers software development and
debugging, data analysis and machine learning, security, and finance and
business analysis. The benchmark contains 100 tasks over 100 distinct skill
groups: 30 single-skill, 46 two-skill, and 24 three-skill tasks.

\paragraph{Metrics and Evaluation Protocol.}
For SkillsBench and SkillEval, we report the mean normalized reward over all
tasks. MolBench-Bind reports exact-match accuracy, while AgentSkillOS-Bench
reports the mean normalized artifact-based score. We express every metric on a
0--100 scale, with higher values indicating better performance. Unless a
no-skill control is explicitly stated, each task is evaluated with its
designated external skills available. Each task is executed in a fresh,
isolated session and workspace, with no model context or task-generated state
carried over between tasks. We run every complete benchmark four times with temperature=0.7 for each
model--harness condition and report the arithmetic mean and sample standard
deviation (with \(n-1\) in the denominator) of the four run-level scores.

\subsection{Overall Effectiveness Across Models and Harnesses}
\label{subsec_overall_effectiveness}

We first examine whether training on verified trajectories consistently
improves skill-augmented agents when the training and evaluation harnesses are
matched. For each backbone--harness combination, \emph{Original} denotes the
off-the-shelf checkpoint, whereas \method{} denotes the same backbone
fine-tuned on verified trajectories collected with the corresponding
evaluation harness. Both checkpoints receive exactly the same task-designated
skills at inference time. This paired comparison keeps the backbone, agent
harness, benchmark, and inference-time skill access fixed, thereby directly
measuring the effect of verified trajectory training in the aligned setting.

Table~\ref{tab:overall_effectiveness} reports results for two backbones, two
agent harnesses, and four benchmarks, yielding
\(2\times2\times4=16\) comparisons. \method{} improves the mean score in every
comparison, with absolute gains ranging from 3.20 to 18.91 points. The smallest
gain occurs for Gemma~4 E4B-IT with OpenCode on SkillsBench, where performance
increases from 7.08 to 10.28. The largest occurs for Qwen3.5-9B with
DeepAgents on SkillEval, where the score rises from 51.62 to 70.53. Thus, the
overall result is not driven by a particular backbone or runtime: all eight
Qwen3.5-9B comparisons, all eight Gemma~4 E4B-IT comparisons, and every
comparison under both OpenCode and DeepAgents improve after training.

\begin{table}[H]
\centering
\small
\begingroup
\renewcommand{\arraystretch}{1.05}
\resizebox{\linewidth}{!}{
\begin{tabular}{lllcccc}
\toprule
\multirow[c]{2}{*}{\textbf{Model}}&\multirow[c]{2}{*}{\textbf{Harness}} &\multirow[c]{2}{*}{\textbf{Checkpoint}} &
\multicolumn{2}{c}{\textbf{General-Purpose Skill Use}}
& \multicolumn{1}{c}{\textbf{Molecular Science}}
& \multicolumn{1}{c}{\textbf{Artifact Production}} \\
\cmidrule(lr){4-5}
\cmidrule(lr){6-6}
\cmidrule(lr){7-7}
 &  &
& SkillsBench & SkillEval
& MolBench-Bind & AgentSkillOS \\
\midrule

\multirow[c]{4}{*}{Qwen3.5-9B}
& \multirow[c]{2}{*}{OpenCode}
& Original
& 5.80 $\pm$ 1.03
& 55.24 $\pm$ 0.72
& 33.11 $\pm$ 2.59
& 79.61 $\pm$ 1.64 \\

&
& \method{}
& \textbf{15.79} $\pm$ 1.69
& \textbf{72.48} $\pm$ 1.32
& \textbf{44.59} $\pm$ 1.56
& \textbf{84.70} $\pm$ 2.17 \\

\addlinespace[2pt]

&
\multirow[c]{2}{*}{DeepAgents}
& Original
& 4.94 $\pm$ 0.70
& 51.62 $\pm$ 1.16
& 31.76 $\pm$ 1.35
& 74.03 $\pm$ 2.03 \\

&
& \method{}
& \textbf{13.62} $\pm$ 0.82
& \textbf{70.53} $\pm$ 1.24
& \textbf{47.30} $\pm$ 1.56
& \textbf{79.84} $\pm$ 1.82 \\

\midrule

\multirow[c]{4}{*}{Gemma 4 E4B-IT}
& \multirow[c]{2}{*}{OpenCode}
& Original
& 7.08 $\pm$ 1.33
& 61.01 $\pm$ 2.05
& 34.46 $\pm$ 1.35
& 49.41 $\pm$ 1.28 \\

&
& \method{}
& \textbf{10.28} $\pm$ 1.16
& \textbf{72.85} $\pm$ 1.30
& \textbf{45.27} $\pm$ 1.35
& \textbf{56.74} $\pm$ 0.93 \\

\addlinespace[2pt]

&
\multirow[c]{2}{*}{DeepAgents}
& Original
& 5.07 $\pm$ 0.95
& 56.14 $\pm$ 1.73
& 31.76 $\pm$ 2.59
& 42.23 $\pm$ 0.85 \\

&
& \method{}
& \textbf{9.24} $\pm$ 1.05
& \textbf{66.39} $\pm$ 2.11
& \textbf{41.89} $\pm$ 1.56
& \textbf{47.38} $\pm$ 1.47 \\
\bottomrule
\end{tabular}
}
\endgroup
\caption{Matched-harness performance with externally supplied skills. Values
are mean $\pm$ sample standard deviation over four complete benchmark runs on
a 0--100 scale. Boldface marks the higher mean within each model--harness
pair.}
\label{tab:overall_effectiveness}
\end{table}

The gains also extend across benchmark categories. On SkillsBench, \method{}
improves performance by 3.20--9.99 points across the four model--harness
settings, while the corresponding gains on SkillEval reach 10.25--18.91
points. Improvements on MolBench-Bind are consistently double-digit,
ranging from 10.13 to 15.54 points, demonstrating that the benefit extends to
specialized molecular-science decisions. AgentSkillOS likewise improves by
5.09--7.33 points across all settings. Notably, Qwen3.5-9B with OpenCode rises
from 79.61 to 84.70 on AgentSkillOS despite the already strong Original
checkpoint, showing that \method{} remains beneficial even at a comparatively
high starting score.

\subsection{Understanding the Improvement}\label{subsec_understanding_improvement}

\subsubsection{Dependence on External Skills}

We next test how much of the SFT advantage remains without inference-time
skill access. On SkillsBench and SkillEval, we compare Original and \method{}
under both harnesses with the designated skill files either withheld or
provided; all other settings remain fixed.

\begin{figure}[b]
\centering
\includegraphics[width=\linewidth]{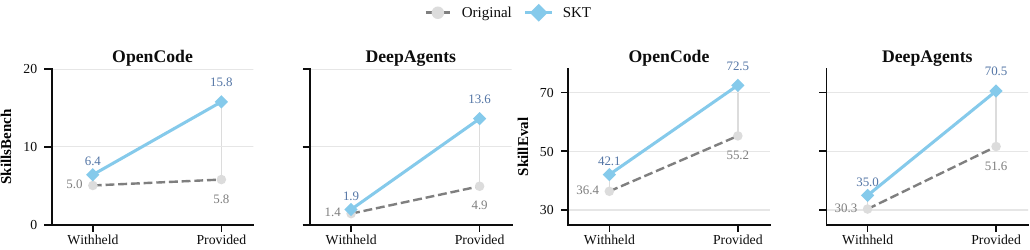}
\caption{Absolute Original and \method{} scores for Qwen3.5-9B with
task-designated skills withheld or provided.}
\label{fig:skill_dependence}
\end{figure}

Figure~\ref{fig:skill_dependence} shows gains of only 0.53--5.69 points when
skills are withheld, versus 8.68--18.91 points when they are provided. This
gap appears on both benchmarks and under both harnesses.

\subsubsection{Verified vs. Unverified Data Pipelines}

We compare the full quality-controlled pipeline against raw synthesis by
constructing a comparable set of tasks and trajectories without verification
or repair, then fine-tuning Qwen3.5-9B with OpenCode on either dataset.

\begin{figure}[t]
\centering
\includegraphics[width=0.68\linewidth]{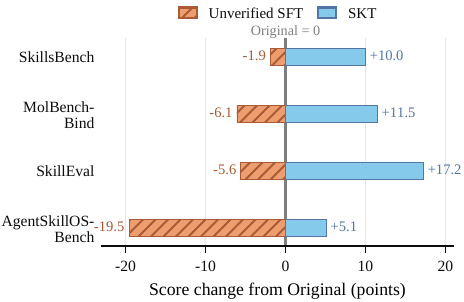}
\caption{Change from Original after Qwen3.5-9B--OpenCode SFT on unverified
or \method{} trajectories (four-run means).}
\label{fig:data_verification}
\end{figure}

Unverified SFT lowers all four benchmark means, while \method{} raises all four
(Figure~\ref{fig:data_verification}); the resulting gaps are 11.91--24.61
points.

\subsection{Harness Transfer and Mixed-Harness Training}
\label{subsec_harness_transfer}

A training trajectory captures not only the task solution and the use of
external skills, but also the interaction protocol imposed by the agent
harness. We therefore ask whether the learned skill-use behavior transfers
across harnesses or remains specific to the environment in which the
trajectories were collected. We study this question through cross-harness
evaluation and mixed-harness training. All evaluations provide the
task-designated skills.

\subsubsection{Cross-Harness Transfer}

We evaluate Qwen3.5-9B on SkillsBench and SkillEval under both OpenCode and
DeepAgents. For each evaluation harness, \emph{Cross-harness SFT} denotes a
checkpoint trained on trajectories collected with the other harness, whereas
\emph{Matched-harness SFT} uses trajectories collected with the same harness
used for evaluation. Original, Cross-harness SFT, and Matched-harness SFT are
therefore evaluated on the same tasks, with the same designated skills and
target-harness configuration; only the source of the SFT trajectories differs.

\begin{figure}[t]
\centering
\includegraphics[width=0.68\linewidth]{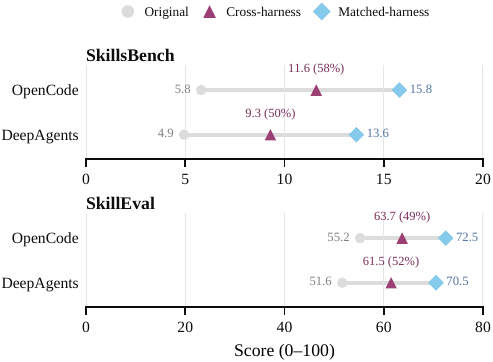}
\caption{Qwen3.5-9B cross-harness transfer. Cross uses the other harness for
SFT; Matched uses the evaluation harness. Labels report retained matched gain
from four-run means.}
\label{fig:cross_harness}
\end{figure}

Cross-harness SFT improves over Original in all four comparisons
(Figure~\ref{fig:cross_harness}). On SkillsBench, transferring from DeepAgents
training to OpenCode evaluation raises the score from 5.80 to 11.60, compared
with 15.79 for Matched-harness SFT, retaining 58.1\% of the matched gain. In
the reverse direction, OpenCode training raises DeepAgents performance from
4.94 to 9.29, compared with 13.62 under matched training, corresponding to
50.1\% retention. The same pattern holds on SkillEval: the transferred
checkpoints retain 49.1\% of the OpenCode matched gain and 52.1\% of the
DeepAgents matched gain.

Across the four settings, Cross-harness SFT yields absolute improvements of
4.35--9.86 points and retains 49.1\%--58.1\% of the corresponding
Matched-harness SFT improvement. Its consistent advantage over Original shows
that a substantial component of the learned skill-use behavior transfers
across agent interfaces. At the same time, Matched-harness SFT remains
stronger in every setting, indicating an additional benefit from aligning the
training trajectories with the target execution environment. This combination
of transferable and harness-aligned gains motivates training a single
checkpoint on trajectories from both harnesses.
\subsubsection{Mixed-Harness Training}

The partial transfer observed above motivates a practical question: can a
single checkpoint support both harnesses without maintaining a separate
specialist for each one? We train Qwen3.5-9B on a mixture of verified OpenCode
and DeepAgents trajectories, and evaluate the resulting \emph{Mixed}
checkpoint under both harnesses. Table~\ref{tab:mixed_harness} compares it with
Original and the corresponding \emph{Specialist}, which is trained only on
trajectories from the evaluation harness.

\begin{table}[H]
\centering
\small
\begingroup
\setlength{\tabcolsep}{4.4pt}
\renewcommand{\arraystretch}{1.08}
\begin{tabular*}{\linewidth}{@{\extracolsep{\fill}}lcccccc@{}}
\toprule
\multirow[c]{2}{*}{\textbf{Benchmark}}& \multicolumn{3}{c}{\textbf{OpenCode}}
& \multicolumn{3}{c}{\textbf{DeepAgents}} \\
\cmidrule(lr){2-4}\cmidrule(lr){5-7}
 & Original & Specialist & Mixed & Original & Specialist & Mixed \\
\midrule

SkillsBench
& 5.80 $\pm$ 1.03
& \textbf{15.79} $\pm$ 1.69
& 14.24 $\pm$ 0.64
& 4.94 $\pm$ 0.70
& 13.62 $\pm$ 0.82
& \textbf{13.93} $\pm$ 1.15 \\

MolBench-Bind
& 33.11 $\pm$ 2.59
& 44.59 $\pm$ 1.56
& \textbf{45.27} $\pm$ 2.59
& 31.76 $\pm$ 1.35
& \textbf{47.30} $\pm$ 1.56
& 44.59 $\pm$ 1.56 \\

SkillEval
& 55.24 $\pm$ 0.72
& 72.48 $\pm$ 1.32
& \textbf{74.05} $\pm$ 1.38
& 51.62 $\pm$ 1.16
& \textbf{70.53} $\pm$ 1.24
& 69.96 $\pm$ 1.62 \\

AgentSkillOS
& 79.61 $\pm$ 1.64
& \textbf{84.70} $\pm$ 2.17
& 82.25 $\pm$ 1.49
& 74.03 $\pm$ 2.03
& \textbf{79.84} $\pm$ 1.82
& 77.40 $\pm$ 2.83 \\

\bottomrule
\end{tabular*}
\caption{Mixed-harness training for Qwen3.5-9B. \emph{Original} is the
off-the-shelf checkpoint, \emph{Specialist} uses trajectories from the
evaluation harness, and \emph{Mixed} uses trajectories from both harnesses.
Scores report mean $\pm$ sample standard deviation over four complete runs;
bold marks the highest mean within each evaluation-harness group.}
\label{tab:mixed_harness}
\endgroup
\end{table}

The Mixed checkpoint improves over Original in all eight
benchmark--harness comparisons, with gains ranging from 2.64 to 18.81 points.
The largest improvements appear on SkillEval, where Mixed raises the OpenCode
score from 55.24 to 74.05 and the DeepAgents score from 51.62 to 69.96.
Substantial gains also hold on SkillsBench and MolBench-Bind under both
harnesses.

Mixed remains close to the matched Specialist throughout the table~\ref{tab:mixed_harness}: their mean
scores differ by at most 2.71 points. Mixed exceeds the Specialist in three
comparisons---by 0.68 points on MolBench-Bind and 1.57 points on SkillEval
under OpenCode, and by 0.31 points on SkillsBench under DeepAgents. In the
remaining five comparisons, the gap is limited to 0.57--2.71 points.

These results show that joint training largely recovers the benefit of
harness-specific alignment while producing one checkpoint that can operate
effectively with either harness. Together with the cross-harness results, this
suggests that shared skill-use behavior and harness-specific interaction
patterns can be learned within a unified model, reducing the need to train and
maintain a separate checkpoint for each execution environment.
\subsection{Scaling the Synthetic Training Pool}
\label{subsec_scaling_skill_coverage}

We jointly scale the synthetic pool from 100 to 2,000 skills and its
corresponding verified OpenCode trajectories, train Qwen3.5-9B under the same
recipe, and evaluate on SkillEval with its designated skills. Original is the
checkpoint at zero training-skill budget. 

\begin{figure}[t]
\centering
\includegraphics[width=0.68\linewidth]{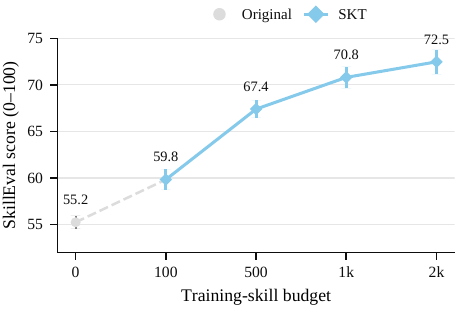}
\caption{SkillEval by training-skill budget for Qwen3.5-9B--OpenCode
(four-run mean $\pm$ standard deviation; equal x spacing).}
\label{fig:scaling_curve}
\end{figure}

SkillEval rises monotonically from 55.24 for Original to 72.48 at 2,000
training skills (Figure~\ref{fig:scaling_curve}).

\subsection{Performance by Skill Cardinality}
\label{subsec_multiskill_analysis}

Finally, we examine whether \method{} remains effective across different
numbers of supplied skills. We partition the 77 SkillsBench tasks into three
groups according to the number \(K\) of task-designated skill artifacts:
\(K=1\) (\(n=25\)), \(K=2\) (\(n=19\)), and \(K\geq3\) (\(n=33\)). Both
checkpoints use Qwen3.5-9B with OpenCode and receive exactly the same
designated skills for each task.

\begin{figure}[t]
\centering
\includegraphics[width=0.68\linewidth]{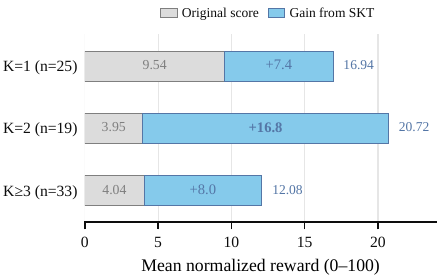}
\caption{SkillsBench reward by skill count for Qwen3.5-9B--OpenCode. Gray is
Original; blue is the \method{} gain. Labels give task counts and final scores.}
\label{fig:multiskill_composition}
\end{figure}

As shown in Figure~\ref{fig:multiskill_composition}, Original obtains mean
normalized rewards of 9.54, 3.95, and 4.04 for \(K=1\), \(K=2\), and
\(K\geq3\), respectively. \method{} increases these scores to 16.94, 20.72,
and 12.08, corresponding to absolute gains of 7.40, 16.77, and 8.04 points.
The improvement is present across all three skill-cardinality groups and is
particularly pronounced for \(K=2\).

Taken together, this breakdown shows that the aggregate benefit of \method{} is
not concentrated in single-skill tasks. Substantial improvements also appear
when multiple skills are supplied, including an 8.04-point gain for tasks with
three or more skills. These results support the effectiveness of \method{} in
strengthening the model's ability to leverage task-provided guidance across
different skill-composition settings.

\subsection{Summary of Findings}

Our experiments demonstrate that SFT on \method{} trajectories improves model performance across diverse skill-use benchmarks. Instead of enabling the model to internalize skills, \method{} enhances its capability to leverage externally supplied skills. Additionally, data ablation studies reveal that these performance gains hinge on training data quality. Directly employing raw synthetic training data without pipeline verification and repair can instead impair skill utilization.

While the training data and corresponding performance improvements are harness-specific, the learned benefits generalize across different harnesses. Training on mixed harness data furthermore yields a single checkpoint compatible with multiple harnesses. We additionally observe that skill-use performance continues to improve as the number of training skills increases.

\section{Conclusion}

We presented \method{}, a three-stage pipeline for creating reliable skill-use
tasks and producing verified execution trajectories for training LLMs to use Agent Skills.
By combining skill selection, task synthesis with feedback-guided repair,
and trajectory synthesis with skill-use verification, \method{} produces
executable examples spanning single- and multi-skill workflows. Across diverse
models, agent harnesses, and application domains, \method{} consistently
improves skill-use performance. The learned behavior also transfers across
harnesses, while mixed-harness training enables a single checkpoint to perform
effectively across different harnesses. Performance further improves as
synthetic skill coverage expands. The results establish verified,
composition-aware data synthesis as a practical route toward more reliable
skill-using agents.

\clearpage
\bibliography{references}

\clearpage
\appendix
\providecommand{\method}{\textsc{SKT}}

\section{End-to-End \method{} Procedure}
\label{app:complete_algorithm}

The main paper presents the three stages of \method{} separately.  Here we
formalize their control flow, including candidate resampling, task repair,
trajectory retry, and the conditions under which a candidate leaves the
pipeline.  The task and trajectory verifiers are treated as abstract
operations; this section specifies when they are invoked and how their outputs
affect the pipeline.

\subsection{Inputs and Outputs}

Let \(\mathcal{C}\) be a collection of externally represented Agent Skills,
\(\mathcal{K}\) the requested skill cardinalities, and \(\mathcal{P}\) the
configured teacher--harness pairs.  A skill configuration
\(\mathcal{S}\subseteq\mathcal{C}\) contains \(k\) distinct skills for some
\(k\in\mathcal{K}\).  The pipeline first constructs a set of accepted
task--configuration pairs \(\mathcal{D}_{\mathrm{task}}\), then collects their
verified execution trajectories \(\mathcal{D}_{\mathrm{traj}}\).  Each retained
trajectory preserves its task, skill, teacher, and harness provenance.

For each \(k\), \(Q_k\) is the target number of accepted tasks and
\(B_{\mathrm{select},k}\) bounds the number of candidate configurations
examined.  \(B_{\mathrm{task}}\) is the maximum number of repairs after initial
task generation, and \(B_{\mathrm{traj}}\) is the maximum number of fresh
rollouts for one task--teacher--harness pairing.

\subsection{Algorithm}

Algorithm~\ref{alg:skt_complete_appendix} gives the complete control flow.
\textsc{QualifyTask} invokes the rule-based, agent-based, and difficulty gates
in that order and returns the first blocking diagnosis.  The agent-based gate
includes both semantic review and the paired skill-dependence test.
\textsc{QualifyTrajectory} combines the deterministic outcome-and-trace check
with the model-based skill-use check.  A trajectory rejection never changes
the task package, and its replacement is sampled without failure or verifier
feedback.  Appendix~\ref{app:qualification_criteria} expands the acceptance
criteria used by these two operations.

\begin{algorithm}[tb]
\caption{End-to-end construction of verified skill-use trajectories.}
\label{alg:skt_complete_appendix}
\small
\begin{algorithmic}[1]
\REQUIRE Skills \(\mathcal{C}\); cardinalities \(\mathcal{K}\); targets
\(\{Q_k\}\); teacher--harness pairs \(\mathcal{P}\); budgets
\(B_{\mathrm{select},k},B_{\mathrm{task}},B_{\mathrm{traj}}\); difficulty
parameters \(N,\theta_{\mathrm{easy}}\)
\ENSURE Tasks \(\mathcal{D}_{\mathrm{task}}\) and trajectories
\(\mathcal{D}_{\mathrm{traj}}\)
\STATE \(\mathcal{U},\mathcal{D}_{\mathrm{task}},
\mathcal{D}_{\mathrm{traj}}\leftarrow\emptyset\)
\FORALL{\(s\in\mathcal{C}\)}
    \IF{\(\textsc{Suitable}(s)\)}
        \STATE \(\mathcal{U}\leftarrow\mathcal{U}\cup\{s\}\)
    \ENDIF
\ENDFOR
\FORALL{\(k\in\mathcal{K}\)}
    \STATE \(n_k,c\leftarrow 0\)
    \WHILE{\(n_k<Q_k\) and \(c<B_{\mathrm{select},k}\)}
        \STATE \(c\leftarrow c+1\); sample \(k\) distinct skills
        \(\mathcal{S}\) from \(\mathcal{U}\)
        \IF{\(k=1\) or \(\textsc{Composable}(\mathcal{S})\)}
            \STATE \(\mathcal{T}\leftarrow\textsc{TaskGen}(\mathcal{S})\)
            \FOR{\(b=0,\ldots,B_{\mathrm{task}}\)}
                \STATE \((v,f)\leftarrow\textsc{QualifyTask}
                (\mathcal{T},\mathcal{S},N,\theta_{\mathrm{easy}})\)
                \IF{\(v=\mathrm{pass}\)}
                    \STATE Add \((\mathcal{T},\mathcal{S})\) to
                    \(\mathcal{D}_{\mathrm{task}}\); \(n_k\leftarrow n_k+1\)
                    \STATE \textbf{break}
                \ELSIF{\(b<B_{\mathrm{task}}\)}
                    \STATE \(\mathcal{T}\leftarrow
                    \textsc{TaskRepair}(\mathcal{T},\mathcal{S};f)\)
                \ENDIF
            \ENDFOR
        \ENDIF
    \ENDWHILE
\ENDFOR
\FORALL{\((\mathcal{T},\mathcal{S})\in\mathcal{D}_{\mathrm{task}}\)}
    \FORALL{\((m,h)\in\mathcal{P}\)}
        \FOR{\(a=1,\ldots,B_{\mathrm{traj}}\)}
            \STATE \(\tau\leftarrow\textsc{Rollout}
            (\mathcal{T},\mathcal{S},m,h)\)
            \IF{\(\textsc{QualifyTrajectory}
            (\tau,\mathcal{T},\mathcal{S})\)}
                \STATE Add \((\mathcal{T},\mathcal{S},m,h,\tau)\) to
                \(\mathcal{D}_{\mathrm{traj}}\); \textbf{break}
            \ENDIF
        \ENDFOR
    \ENDFOR
\ENDFOR
\RETURN \(\mathcal{D}_{\mathrm{task}},\mathcal{D}_{\mathrm{traj}}\)
\end{algorithmic}
\end{algorithm}

\subsection{Feedback and Stopping Policy}

Algorithm~\ref{alg:skt_complete_appendix} uses feedback only for task repair: the first
blocking task diagnosis is returned to \textsc{TaskRepair}, whereas a rejected
trajectory triggers a fresh retry without verifier feedback.  The finite
repair and rollout budgets define when each branch stops.  Appendix~
\ref{app:qualification_criteria} specifies the corresponding acceptance
evidence and retention rules.

\subsection{High-Level Experimental Configuration}

Table~\ref{tab:skt_algorithm_configuration} records the configuration needed
to interpret Algorithm~\ref{alg:skt_complete_appendix}.  Training, inference, hardware,
serialization, and software details are reported separately with the
experimental reproducibility settings.

\begin{table}[tb]
\centering
\begin{tabular}{@{}p{0.35\columnwidth}p{0.57\columnwidth}@{}}
\toprule
\raggedright\textbf{Component} &
\raggedright\textbf{Experimental setting} \tabularnewline
\midrule
\raggedright Skill cardinalities &
\raggedright \(\mathcal{K}=\{1,2,3\}\) \tabularnewline
\raggedright Task-side model components &
\raggedright DeepSeek V4 Pro with a Claude Agent harness \tabularnewline
\raggedright Difficulty controller &
\raggedright Qwen3.5-35B-A3B with OpenCode \tabularnewline
\raggedright Difficulty parameters &
\raggedright \(N=5\), \(\theta_{\mathrm{easy}}=0.6\) \tabularnewline
\raggedright Teacher models &
\raggedright MiniMax-M2.5, GLM-5, Qwen3.5-397B-A17B, and DeepSeek V4 Pro
\tabularnewline
\raggedright Trajectory harnesses &
\raggedright DeepAgents and OpenCode \tabularnewline
\bottomrule
\end{tabular}
\caption{High-level configuration used for \method{} data construction.}
\label{tab:skt_algorithm_configuration}
\end{table}

\section{Detailed Qualification Criteria}
\label{app:qualification_criteria}

Algorithm~\ref{alg:skt_complete_appendix} leaves \textsc{QualifyTask} and
\textsc{QualifyTrajectory} abstract.  This section specifies their decision
boundaries and recorded evidence.  We report the verifier interfaces and pass
conditions rather than reproducing prompt templates or implementation source
code.  The resulting specification is independent of a particular authoring
model or agent harness.

\subsection{Verifier Inputs and Hidden Assets}

For a task package \(\mathcal{T}\), let \(V(\mathcal{T})\) contain the
solver-visible instruction, setup artifacts, execution settings, and output
contract.  The selected skills \(\mathcal{S}\) are mounted separately.  Hidden
assets \(H(\mathcal{T})\) contain the evaluator specification, executable reference solution, and task-authoring metadata that
records the intended role of each skill.  A solver receives
\(V(\mathcal{T})\) and, in the with-skill condition, \(\mathcal{S}\), but never
receives \(H(\mathcal{T})\).  Task verifiers may inspect both visible and
hidden surfaces; trajectory verifiers additionally inspect the realized tool
trace, output artifacts, and evaluator outcome.  In the paired skill dependence
test, only access to \(\mathcal{S}\) changes between the two solver runs.

\subsection{Task-Side Qualification}

For the reported corpus of 4,000 accepted task packages, acceptance requires
passing the same three top-level gates in sequence: rule-based verification,
agent-based verification, and difficulty control.  The agent-based gate
comprises both semantic task review and the paired skill-dependence test.
Table~\ref{tab:task_qualification} expands the conditions used by
\textsc{QualifyTask}; every listed row is mandatory.

\begin{table*}[t]
\centering
\small
\begin{tabular}{@{}p{0.20\textwidth}p{0.76\textwidth}@{}}
\toprule
\raggedright\textbf{Check} &
\raggedright\textbf{Pass condition} \tabularnewline
\midrule
\raggedright Rule-based: package and execution &
\raggedright The package is complete and parseable; paths are valid; evaluator
definitions are supported; and the authoring trace records access to every
skill in \(\mathcal{S}\).  In a clean workspace initialized only from task
inputs, the reference solution executes successfully and its produced artifacts
receive full evaluator credit.  Visible task materials do not copy reference
values or required skill rules verbatim.
\tabularnewline
\raggedright Agent-based: semantic validity &
\raggedright The instruction clearly identifies the required artifact and
output location; the local environment contains all information needed for
completion; and solver-visible materials outside the supplied skills do not
semantically reveal the hidden answer, evaluation logic, or decisive
skill-derived requirements. \tabularnewline
\raggedright Agent-based: skill dependence &
\raggedright Paired with-skill and without-skill rollouts hold the task,
solver, harness, and all remaining execution settings fixed.  Writing their
evaluator scores as \(r^{+}\) and \(r^{-}\), respectively, the scores must
satisfy \(r^{+}>r^{-}\), establishing that skill access benefits task completion
at the bundle level.  This paired test does not require independent
leave-one-skill-out necessity for each member.
\tabularnewline
\raggedright Difficulty control &
\raggedright With the selected skills available, a fixed solver and harness
attempt the task for \(N\) independent rollouts.  If \(r_j\) is the evaluator
score, the task passes only when
\(p_{\mathrm{pass}}(\mathcal{T},\mathcal{S})=
N^{-1}\sum_{j=1}^{N}\mathbf{1}[r_j=1]<\theta_{\mathrm{easy}}\).  We use \(N=5\) and
\(\theta_{\mathrm{easy}}=0.6\); hence three or more full-credit outcomes mark
the task as too easy and return it for repair. \tabularnewline
\bottomrule
\end{tabular}
\caption{Task-side qualification checks and pass conditions.}
\label{tab:task_qualification}
\end{table*}

\subsection{Feedback-Guided Repair}

Qualification stops at the first failing top-level gate.  Its structured
diagnosis contains concrete package errors, semantic feedback, or
trajectory-grounded difficulty suggestions.  \textsc{TaskRepair} revises the
complete task package rather than patching a solver trajectory.  The revised
package then restarts qualification at the rule-based gate, because any change
can invalidate earlier evidence.  A task enters
\(\mathcal{D}_{\mathrm{task}}\) only after one package version passes all three
gates; a candidate that exhausts its repair budget is discarded.

\subsection{Trajectory-Side Qualification}

Trajectory qualification consists of a deterministic validator followed by an
LLM validator.  The deterministic validator uses harness-appropriate terminal
signals but applies the same four requirements in both harnesses: full
evaluator credit, normal and complete termination, a well-formed tool trace,
and explicit access to the required skills.  An execution failure, incomplete
termination, malformed trace, or non-full evaluator outcome is blocking.

The LLM validator examines the complete realized trace and checks every
designated skill individually.  Each skill must be consulted before the action
it should guide, affect a concrete decision or operation, and be applied
correctly.  Evidence may include a skill-derived convention, formula, API
pattern, threshold, ordering constraint, resource, verification step, or
repair that changes the produced artifact or workflow.  Merely opening or
paraphrasing a skill file, citing it after the relevant action, or completing
the task entirely through generic reasoning is insufficient.  For a
multi-skill task, all designated skills must make distinct substantive
contributions within one coherent workflow.  A trajectory passes only if both
validators pass.

\subsection{Retry, Retention, and Recorded Evidence}

A rejected rollout is discarded.  The next attempt for the same
task--teacher--harness pairing begins in a fresh session and workspace and does
not receive the previous trace, evaluator outcome, failure reason, or verifier
feedback.  The first rollout that passes both validators is retained; if the
rollout budget is exhausted, that pairing contributes no training example.
The retained example contains the uninterrupted solver interaction, while
hidden references, evaluator internals, and verifier analyses remain outside
the solver trace.

For each task attempt, the verification record stores the gate verdict, a
concise failure category, and the associated task and skill identifiers.  For
each trajectory attempt, it stores the teacher and harness identifiers,
evaluator outcome, terminal status, validator verdicts, and the skill-use
evidence used for the decision.  These fields support aggregate auditing of the
reported criteria without exposing hidden task assets, verifier prompts, or
implementation source code.  Token-length filtering, serialization, and loss
masking are training-data conversion decisions and are specified separately in
the SFT reproducibility section.

\section{Dataset Composition and Trajectory Statistics}
\label{app:dataset_statistics}
\setcounter{dbltopnumber}{2}

This appendix reports aggregate properties of the dataset used in the main
experiments.  The counts follow the same accepted task pool as the main
experimental setup: 4,000 task packages built from 2,000 skills selected from
\texttt{skills.sh} and paired with four teachers and two execution harnesses.

\subsection{Task and Skill Composition}

Table~\ref{tab:accepted_task_composition} breaks down the accepted task pool by
skill cardinality.  Each accepted task is eligible for eight
trajectory-collection pairings, one for
each teacher--harness combination, yielding 32,000 candidate pairings before
trajectory-side verification.

\begin{table}[H]
\centering
\small
\setlength{\tabcolsep}{3pt}
\begin{tabular}{@{}lrrr@{}}
\toprule
\textbf{\(k\)} & \textbf{Tasks} &
\textbf{Skill refs.} & \textbf{Pairings} \\
\midrule
\(k=1\) & 1,520 & 1,520 & 12,160 \\
\(k=2\) & 1,295 & 2,590 & 10,360 \\
\(k=3\) & 1,185 & 3,555 & 9,480 \\
\midrule
Total & 4,000 & 7,665 & 32,000 \\
\bottomrule
\end{tabular}
\caption{Accepted task composition by skill cardinality.  The \emph{Skill
refs.} column counts designated task--skill associations.}
\label{tab:accepted_task_composition}
\end{table}

Table~\ref{tab:selected_skill_taxonomy} gives
an illustrative, overlapping organization of functions represented in the
selected pool, based on package names and descriptions and, where available,
optional metadata.  For example, generating a statistical report can combine
data analysis with document creation.

\begin{table*}[t]
\centering
\small
\begin{tabular}{@{}p{0.20\textwidth}p{0.39\textwidth}p{0.34\textwidth}@{}}
\toprule
\textbf{Functional family} & \textbf{Scope} &
\textbf{Representative skill families} \\
\midrule
Software engineering and quality
& Developing, testing, debugging, reviewing, and maintaining software.
& Code review; test repair; API and framework implementation; Git and
repository workflows. \\
Infrastructure, systems, and security
& Deploying, operating, monitoring, and securing systems and services.
& CI/CD; containers and Kubernetes; cloud and observability; vulnerability
assessment. \\
Data, ML, and scientific computing
& Transforming structured data, building learned systems, and performing
quantitative or scientific computation.
& ETL and SQL; forecasting and statistics; computer vision; scientific
analysis. \\
Documents, content, and media
& Creating, transforming, inspecting, or validating human-facing artifacts.
& PDFs, spreadsheets, and presentations; technical writing; UI design; image,
audio, and video processing. \\
Business, finance, and operations
& Supporting organizational decisions and recurring professional workflows.
& Market and financial analysis; CRM and sales; product analysis and planning;
project and people operations. \\
Research, communication, and productivity
& Gathering, synthesizing, communicating, and organizing information.
& Literature and citation workflows; deep research; email, calendar, and
knowledge organization. \\
Agent tooling and workflow automation
& Building tool-using agents and automating interactions with external
services.
& MCP and tool construction; prompt, context, and memory workflows;
multi-agent and browser automation. \\
\bottomrule
\end{tabular}
\caption{An overlapping functional taxonomy of the selected 2,000-skill pool.
The examples are illustrative rather than exhaustive labels; skills and tasks
are not forced into mutually exclusive categories.}
\label{tab:selected_skill_taxonomy}
\end{table*}

Because these families overlap, we keep them separate from the compositionality
statistics above.  The \(k=1,2,3\) tiers count designated skill artifacts, not
functional domains.  A multi-skill task may combine skills within one family
or across several families.

\subsection{SkillEval Benchmark Composition}
\label{app:skilleval_composition}

We construct SkillEval with the same task-synthesis pipeline used for the
training task pool, but from a separately generated candidate pool.  SkillEval
is reserved for evaluation: neither its 100 task packages nor trajectories
collected by executing them are included in the SFT data.  The resulting
separation is therefore at the task-package and trajectory levels.

Table~\ref{tab:training_skilleval_composition} compares the skill-cardinality
composition of the two task pools.  Each of SkillEval's 100 tasks is associated
with one designated skill group.  These groups are distinct from one another
within SkillEval and are drawn from a held-out skill pool disjoint from the
training skill pool.  Of the 100 tasks, 30 are single-skill, 46 are two-skill,
and 24 are three-skill.

\begin{table}[tb]
\centering
\small
\setlength{\tabcolsep}{4pt}
\begin{tabular}{@{}lrrrr@{}}
\toprule
\textbf{Task pool} & \textbf{Tasks} & \textbf{\(k=1\)} &
\textbf{\(k=2\)} & \textbf{\(k=3\)} \\
\midrule
Training & 4,000 & 1,520 & 1,295 & 1,185 \\
SkillEval & 100 & 30 & 46 & 24 \\
\bottomrule
\end{tabular}
\caption{Skill-cardinality composition of the training task pool and
SkillEval.}
\label{tab:training_skilleval_composition}
\end{table}

The benchmark spans several functional areas, including software development
and debugging, data analysis and machine learning, security, and finance and
business analysis.  During evaluation, each task is executed in a fresh
isolated session and workspace with its designated external skills available,
except in the explicit no-skill controls.  We report mean normalized reward
over all 100 tasks.  For each model--harness condition, we run the complete
benchmark four times, following the protocol in the main experimental setup.

\begin{table}[t]
\centering
\small
\setlength{\tabcolsep}{2pt}
\begin{tabular*}{\columnwidth}{@{\extracolsep{\fill}}lrrrr@{}}
\toprule
\textbf{Harness} & \textbf{\(N\)} &
\textbf{\shortstack{Assistant\\turns}} &
\textbf{\shortstack{Tool\\calls}} &
\textbf{\shortstack{Skill-load\\requests}} \\
\midrule
DeepAgents & 14,277 & 12 / 17 & 18 / 26 & 5 / 9 \\
OpenCode   & 12,887 & 9 / 13  & 11 / 16 & 3 / 7 \\
\bottomrule
\end{tabular*}
\caption{Descriptive statistics of the collection-level retained trajectories.
Interaction-count entries report median / 90th percentile.}
\label{tab:retained_trajectory_statistics}
\end{table}

\subsection{Trajectory Measurements}

We summarize the retained verified trajectory pools at the collection level.
The \(N\) values therefore match the trajectory counts in the main experimental
setup rather than denoting a per-backbone count after target-specific
rendering.  Table~\ref{tab:retained_trajectory_statistics} reports the median
and 90th percentile for each interaction-count measure.

We count each assistant-role message as one assistant turn and each function
invocation as one tool call; parallel calls count separately.  Skill-load
requests are native \texttt{skill} calls for OpenCode and \texttt{read\_file}
calls targeting \texttt{/skills/.../SKILL.md} for DeepAgents.  Repeated requests
and zeros remain in the distributions.  These syntactic counts do not
themselves establish successful skill loading, substantive skill use, or
trajectory acceptance.

\section{Training and Evaluation Details}
\label{app:training_evaluation_details}

We report data-conversion, optimization, and evaluation details for Qwen3.5-9B
and Gemma~4 E4B-IT.  The tokenizer and chat template are model-specific, while
the tool representation remains harness-native.

\subsection{Trajectory Serialization and Supervision}

Each retained rollout is serialized end to end while preserving its harness's
native tool schema and action structure.  Post-termination acknowledgements and
logging events are excluded, and the two harnesses are not mapped to a common
tool vocabulary.  After rendering with the target tokenizer and chat template,
we supervise model-authored reasoning, tool calls, and responses, while masking
system and task messages and tool observations as in the main SFT objective.
We use neither history masking nor packing and discard, rather than truncate,
records longer than 64k tokens.

\subsection{Optimization and Checkpoint Selection}

We perform full-parameter SFT with LLaMA-Factory and DeepSpeed ZeRO-3, using the
settings in Table~\ref{tab:sft_implementation}.  Records are shuffled with a fixed seed;
runs neither resume from optimizer state nor use validation or benchmark scores
for selection.  Evaluation loads the final checkpoint after one epoch.

\begin{table}[H]
\centering
\small
\begin{tabular}{@{}p{0.31\columnwidth}p{0.61\columnwidth}@{}}
\toprule
\raggedright\textbf{Setting} &
\raggedright\textbf{Value} \tabularnewline
\midrule
\raggedright Epochs &
\raggedright 1 \tabularnewline
\raggedright Optimizer &
\raggedright Fused AdamW; \(\beta_1=0.9\), \(\beta_2=0.999\),
\(\epsilon=10^{-8}\), weight decay \(=0\) \tabularnewline
\raggedright Learning rate &
\raggedright \(5\times10^{-6}\); cosine decay with a 3\% warmup ratio
\tabularnewline
\raggedright Global batch &
\raggedright 8 (8 workers \(\times\) 1 example per worker \(\times\) 1
gradient-accumulation step) \tabularnewline
\raggedright Precision and memory &
\raggedright bfloat16; gradient checkpointing; ZeRO-3; label-only fused
cross-entropy \tabularnewline
\raggedright Gradient clipping &
\raggedright Global norm 1.0 \tabularnewline
\raggedright Sequence handling &
\raggedright No packing or token-level truncation; a whole-record limit of
64k tokens \tabularnewline
\raggedright Checkpoint rule &
\raggedright Final one-epoch checkpoint; no validation-based selection
\tabularnewline
\bottomrule
\end{tabular}
\caption{Common SFT implementation settings.}
\label{tab:sft_implementation}
\end{table}

\begin{figure}[b]
  \centering
  \includegraphics[width=0.98\textwidth]{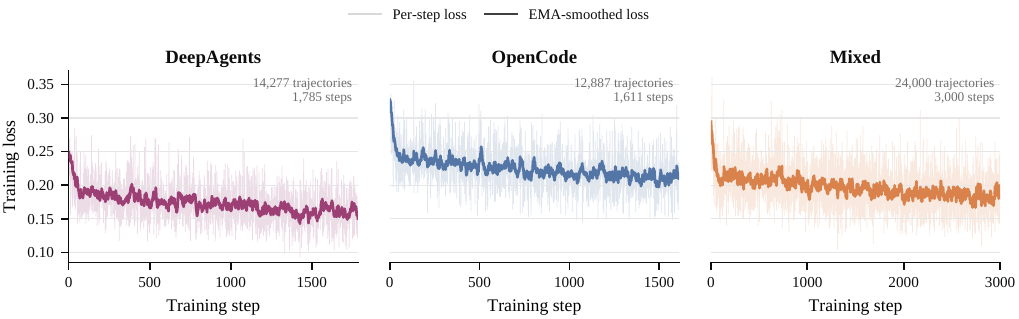}
  \caption{Qwen3.5-9B training-loss curves for the DeepAgents, OpenCode, and
  mixed-harness SFT runs. Thin lines show per-step loss and thick lines show
  the exponential moving average. Labels give the number of training
  trajectories and optimizer steps in each run.}
  \label{fig:training_curves}
\end{figure}

Figure~\ref{fig:training_curves} reports the corresponding one-epoch
optimization traces for Qwen3.5-9B.  The per-step losses are noisy, while the
EMA-smoothed curves decrease across the DeepAgents, OpenCode, and balanced
mixed-harness runs.

The environment uses Transformers~5.6.0, PyTorch~2.10.0 with CUDA~12.8,
Datasets~4.0.0, and Tokenizers~0.22.2; eight GPU workers implement the reported
effective batch size.

\subsection{Inference and Evaluation Protocol}

Models are served through OpenAI-compatible endpoints and invoked through
harness-native loops.  DeepAgents reads materialized
\texttt{/skills/.../SKILL.md}
files, whereas OpenCode uses its native \texttt{skill} tool.  Formal evaluation
uses temperature 0.7; other decoding settings are fixed, while timeout
watchdogs and interaction limits remain benchmark- and harness-specific.

Original and SFT receive identical task assets and designated skills except in
explicit no-skill controls.  Cross-harness evaluation changes the agent loop
and skill interface, not the checkpoint.  Every task starts in a fresh isolated
session and workspace with no carried context or state.

We use each benchmark's native evaluator and the 0--100 metrics defined in the
main setup.  For each model--harness condition, four complete runs yield four
aggregate scores; we report their arithmetic mean and sample standard deviation
with \(n-1\), rather than computing dispersion over task-level scores.

\subsection{Mixed-Harness and Scaling Variants}

For mixed-harness training, we seed-sample 12,000 trajectories from each
harness's final pool, yielding a balanced set of 24,000 verified records.

We preserve each harness's native message and tool-call representation and
inject no harness identifier into the training messages.
The checkpoint otherwise uses the common SFT settings in
Table~\ref{tab:sft_implementation}.

For scaling, we train Qwen3.5-9B with the common recipe on verified OpenCode
trajectories from 100, 500, 1,000, or 2,000 skills, then evaluate on SkillEval
with designated skills. Zero budget denotes the off-the-shelf Original, not a
separate SFT run.

\end{document}